\documentclass[conference]{IEEEtran}
\IEEEoverridecommandlockouts
\usepackage{cite}
\usepackage{amsmath,amssymb,amsfonts}
\usepackage{algorithmic}
\usepackage{graphicx}
\usepackage{textcomp}
\usepackage{xcolor}
\def\BibTeX{{\rm B\kern-.05em{\sc i\kern-.025em b}\kern-.08em
    T\kern-.1667em\lower.7ex\hbox{E}\kern-.125emX}}
\begin{document}

\title{The speaker-independent lipreading play-off; \\ a survey of lipreading machines
}

\author{\IEEEauthorblockN{Jake Burton$^1$, David Frank$^1$, Mahdi Saleh$^1$, Nassir Navab$^1$, and Helen L. Bear$^{1,2}$}
\IEEEauthorblockA{\textit{$^1$Computer Aided Medical Procedures and Augmented Reality, Institute for Informatics, Technical University Munich Germany} \\
\textit{$^2$School of Electrical Engineering and Computer Science, Queen Mary University of London, United Kingdom}\\
\{j.burton, d.frank, m.saleh, n.navab, dr.bear\}@tum.de}
}


\maketitle

\begin{abstract}
Lipreading is a difficult gesture classification task. One problem in computer lipreading is \textit{speaker-independence}. Speaker-independence means to achieve the same accuracy on test speakers not included in the training set as speakers within the training set. Current literature is limited on speaker-independent lipreading, the few independent test speaker accuracy scores are usually aggregated within dependent test speaker accuracies for an averaged performance. This leads to unclear independent results. 
Here we undertake a systematic survey of experiments with the TCD-TIMIT dataset using both conventional approaches and deep learning methods to provide a series of wholly speaker-independent benchmarks and show that the best speaker-independent machine scores 69.58\% accuracy with CNN features and an SVM classifier. This is less than state-of-the-art speaker-dependent lipreading machines, but greater than previously reported in independence experiments.
\end{abstract}


\section{Introduction}
Lipreading machines are developed using expertise from computer vision and speech processing. Such systems undertake face analysis for small gesture recognition from the lips \cite{stafylakis2017combining}. Lipreading recognition is the classification of units within a visual speech signal \cite{bear2014resolution}. Visemes, a term used broadly for the `units' of visual speech, are not formally defined, and there is no agreed dictionary of such units \cite{bear2017speaker}. This means visual speech recognition (also known as lipreading) is difficult. Lipreading is made all the more a challenging problem because the lips are a small deformable surface with no skeletal structure and whilst recognition of visual speech is aided by the visibility of the teeth and tongue \cite{lan2010improving}, contrast over this small area of the face is limited which makes discrimination of small lip motions difficult. 

Machine lipreading (MLR) is a niche area of computer vision and speech processing. The concept of a lipreading machine is exciting not only because of the fame from science-fiction folklore, but because of the many potential benefits to society. MLR has many possible uses for example; in aviation, law enforcement and security \cite{bowden2013recent}, for sports refereeing \cite{zidane}, understanding historical documentaries and silent films \cite{silentmovie}, and in health care \cite{witte2000elderly}. These are in addition to its major potential as an adjunct in audio-visual speech recognition (AVSR) systems (e.g. \cite{petridis2018audio}).

It is unlikely that the final end users of such a system will be part of the training corpus so resolving speaker-independence is important for future usability. Most research in MLR is either; totally speaker-dependent -- where speakers are both test and training or, speaker-independent classifications are aggregated with speaker-dependent ones, i.e. the set of test speakers is a mixture of both training speakers and non-training speakers.
Table~\ref{tab:sidef} shows an example of speaker-independence based on the definition in \cite{bear2017visual}, it shows that three speakers are used to train, one training speaker (Sp$_2$) and one training-independent speaker (Sp$_4$) is used to validate, (Sp$_2$ is present in both training and validation but the samples used in each phase remain distinct), but importantly, all test speakers (Sp$_5$ and Sp$_6$) are not included in the training or validation. 

\setlength{\tabcolsep}{4pt}
\begin{table}[!h]
\centering
\caption{Speaker splits for speaker-independent test results.}
\label{tab:sidef}
\begin{tabular}{lll}
\hline\noalign{\smallskip}
Training speakers & Validation Speakers & Test Speakers \\
\noalign{\smallskip}
\hline
\noalign{\smallskip}
Sp$_1$ & Sp$_2$ & Sp$_5$ \\
Sp$_2$ & Sp$_4$ & Sp$_6$\\
Sp$_3$ & & \\
\hline
\end{tabular}
\end{table}
\setlength{\tabcolsep}{1.4pt}

Deep learning classification methods have opened up new ways to address speaker-independence (as we discuss in Section~\ref{sec:back}) but as yet we have not seen a systematic and fair comparison of the architectures of conventional lipreading systems and the most representative deep learning approaches. This work is a series of tests to see which lipreading system is the most accurate for speaker-independent lipreading. We observe which system components have the greatest influence on system accuracy.
%
We do not test all variants of possible deep learning architectures for speaker-independent lipreading as there are almost infinite options. Rather this work establishes a set of benchmarks with the most common systems of deep learning and conventional pipelines so that future researchers can expand upon this position paper.  

The rest of this paper is structured as follows. We summarise the latest work contributing to speaker-independent lipreading machines, before describing a `play-off' methodology and survey implementation. A results analysis compares the accuracy of different architectures and before concluding with a set of wholly speaker-independent benchmarks.

\section{BACKGROUND}
\label{sec:back}
Human lipreaders are in reality speechreaders. They use all information in a conversation environment to comprehend what is being said such as; emotions, hand gestures, and prior knowledge of language and environment context \cite{summerfield1992lipreading,bear2017visual}. Skilled human lipreaders have demonstrated robust accuracy is speaker-dependent \cite{yakel2000effects}. Most commonly human lipreaders are fairly inaccurate particularly when lipreading from video \cite{lan2012insights}, but speechreading accuracy increases when a relationship is developed between the communicator and the reader \cite{lott1960influence} i.e. a lipreader can `learn' to read individuals. 

Conversely work in developing a lipreading machine uses only the mouth information. In \cite{assael2016lipnet} and \cite{son2017lip} the authors present work which lipreads whole sentences and the accuracies achieved are $95.2\%$ and $97\%$ respectively on a grammatically very simple dataset (GRID \cite{cooke2006audio}). On a large continuous speech dataset Chung \textit{et al.} achieved $76.2\%$. These accuracies benefit from the discriminatory power of a long sequence of lip gestures. Therefore one approach to address speaker-independence is to only lipread grammatically correct, and long, sentences. In the real world this is impractical as lipreading one-word sentences accurately is just as important as a sentence of five words. Also most people do not talk with perfect grammar. These works do not state if results are speaker-dependent or speaker-independent so given the speaker organization in the datasets used we have to assume these are speaker-dependent results or speaker-dependent results with some independent test speakers.

More MLR literature focused on recognizing words. An early lipreading paper to use a deep learning approach \cite{wand2016lipreading} achieved accuracy of $79.6\%$ with a simple task because of the limited grammar variation in the GRID dataset. Also the small number of speakers ensures this is a speaker-dependent setup. In \cite{Chung2018} the authors construct a complex deep learning architecture to achieve $71.5\%$, but any effects of independent test speakers are grouped in with dependent test speakers and are thus unknown. In these examples wholly speaker-independent results have not been reported. 
 
One approach towards speaker independence with conventional lipreading systems was presented in \cite{bear2017speaker}. Here the authors used speaker-specific visemes \cite{bear2014phoneme} to mitigate the negative effects of speaker-independence without success on isolated words (AVL2) \cite{cox2008challenge}. But in a second experiment using the same approach on continuous speech data \cite{bear2017bmvcVariability} significant recovery of accuracy was achieved, although still not as accurate as speaker-dependent approaches. This outcome is attributed to capturing a datasets language model within each speakers visemes in addition to speaker individuality. 

Sequences of visemes form the visual speech signals that lipreading machines learn to recognise. One suggested reason lipreading machines fail in independent settings is because  these signals are not consistent across speakers. In \cite{bear2018comparing} we learn that the appearance of a speakers lips changes subject to pronunciation, and speakers will have different visemes sequences for consistent sentences thus any independent machine will need to learn all variations. One approach for speaker-independence is to use speaker specific phoneme-to-viseme maps \cite{bear2017phoneme} but these increase the data requirements for lipreading systems and does not solve the independence objective as a phoneme-to-viseme map would be needed for all test speakers. Therefore in these experiments we use the best speaker-independent visemes previously presented (we describe the ground truth preparation in section~\ref{sec:data}). 

The two most significant works in speaker-independent lipreading are by Wand et al.\cite{wand2017improving} and Almajai et al. \cite{almajai2016improved}. The first paper uses a machine learning technique (domain adaptation) and the second, a technique from audio speech processing (maximum likelihood linear regression (MLLR)). In both cases the suggestion is that each speaker has their own visual speech space, and by some function, one can adapt a model from one speaker to another. Both approaches had a positive but not significant, increased word accuracy on independent test speakers. 

\section{METHODOLOGY}
\label{sec:meth}

\subsection{Dataset preparation}
\label{sec:data}
TCD-TIMIT \cite{harte2015tcd} is an audio-visual speech data set that includes 59 volunteer speakers, uttering in total $6913$ phonetically rich sentences in Hibeniro English. The dataset is non-proprietary for research and all speakers are filmed straight on to avoid lip occlusion. We divide the $59$ speakers as follows: $50$ training speakers and nine (randomly selected) test speakers. We repeat this five times for five-fold cross-validation with replacement to assure a speaker-independent testing regime. For CNN and LSTM classifiers eight randomly chosen samples are held out for each of the $50$ training speakers for validation. 

The outputs of our data preparation is labeled lip images. Our data preparation pipeline is: conversion of each video clip into single frames using \texttt{FFmpeg}. These frames are converted to greyscale before face detection using the \texttt{dlib} library and the iBUG 68 landmark predictor \cite{3dmenpo}. We extract only the mouth outline and inner points from the detected facial landmarks. All images are cropped to these landmarks plus a boundary of $10$px before the image is resized to $48\times48$px.  We extract different features as described in Section~\ref{sec:Features}.

%



To prepare ground truths labels for each frame TCD-TIMIT provides a set of time-stamped phoneme labels and these are adapted into per frame labels so every frame has a phoneme label. There is not an equal number of training samples per phoneme class due to the linguistic content of the spoken sentences. To create a viseme ground truth (viseme labels are a combination of Woodward \cite{woodward1960phoneme} and Disney \cite{disney} visemes. These visemes perform best in prior lipreading tests on continuous speech \cite{bear2017phoneme}) the phoneme-to-viseme map translates the phoneme labels (per frame) into visemes and for two ground truths, one for the acoustic space, another for the visual space. 



As discussed in Section~\ref{sec:back}, some prior work report word accuracy (e.g. \cite{bear2016decoding}). This can be achieved with a variety of methods such as; using word labels for classes, using a pronunciation dictionary lookup after phoneme/viseme classification, or using a word-level language model for grammar discrimination \cite{bear2018visual}. Whichever pipeline is selected, the final classification accuracies are affected, usually positively. For example, using a word language model increases the accuracy significantly, compared to building word-labeled classifiers which requires much more data \cite{thangthaicomparing}. However, as our results are to be benchmarks irrespective of the language influence, we report our results as both phoneme and viseme accuracy ($A = \frac{\#correct}{\#total} \cdot 100$) without word decoding\footnote{\texttt{https://github.com/drylbear/visualspeechplayoffs}}.

\section{EXPERIMENTS}

\subsection{Features}
\label{sec:Features}
We do not describe the features in depth as the mathematics are well covered in literature instead we detail our implementation. All features are normalised to have zero mean.

\textbf{Histogram of Oriented Gradients} (HoG) feature vectors  are computed with the \texttt{hog} method from \texttt{scikit-image} with nine orientations, $8\times8$ cells and four cells ($2\times2$) per block. Blocks are normalized using the $L^{2}$-norm and the values thresholded at $0.2$ and then renormalized with \textit{`L2-Hys'} block normalization scheme. 
HoG features are fairly accurate in speaker-dependent lipreading systems e.g. \cite{palevcek2016lipreading} and \cite{pei2013unsupervised}. 

\textbf{Scale-Invariant Feature Transform} (SIFT) features are computed on the ROI using OpenCV \texttt{sift.detectAndCompute}. The resulting $\#keypoints \cdot 128$ matrix of keypoint descriptors are summarized into a single global feature by implementing `bag of keypoints' locality-constrained linear coding with the \texttt{feature-aggregation} library. The `bag of words' model consists of $20$ codewords and is trained on the SIFT features extracted from $20$ random frames for each speaker ($1180$ frames in total). This trained model then transforms the local descriptors into the final feature vectors. SIFT features are a robust feature method and MLR systems have tried augmenting them with other features, \cite{sterpu2018towards}.

\textbf{Discrete Cosine Transform} (DCT) features are produced by computing the $2$D DCT-II of the lip ROI images using the \texttt{dct} method from \texttt{OpenCV}, and then selecting the first ten DCT coefficients (including the DC bias term) in a characteristic zig-zag pattern. This allows us to shorten the feature vectors without losing the coefficients with the highest impact, as in natural images most energy is present in the lower frequencies \cite{guan2012multimedia}.  DCT variations for MLR include 3-D DCT with modified classifiers, e.g. \cite{min2011lip}. 


We implement \textbf{Active Appearance Models} (AAM) with the Menpo library \cite{alabort2014menpo} holistic AAM. This produces a holistic appearance representation with a non-linear warp function for transforming the texture into the reference mesh.
To extract AAM features we fit the AAM using Lukas-Kanade optimization together with the Wiberg Inverse Compositional (WIC) Gauss-Newton algorithm for affine image alignment between adjacent frames (without landmarks). 
The maximum number of iterations permitted for convergence is $80$. For each frame a single feature vector is extracted which is the concatenation of the shape and appearance parameters. There are many examples of speaker-dependent AAMs improving MLR \cite{bear2014resolution}.

In end-to-end CNN classification the architecture includes first convolutional layers learning \textbf{deep image / CNN} features before classification using (commonly) Softmax regression. Here we both use this approach for the CNN classification on image inputs, but we also save the image features to use them as inputs to our conventional classifiers. We describe the first part of our end-to-end system here and the classification later in Section~\ref{sec:classification}.

We use the ResNet50 architecture \cite{2015arXiv151203385H} for our CNN twice. In section~\ref{sec:classification} we describe end-to-end training for classification (e.g. \cite{noda2014lipreading} but first we extract deep features \cite{chung2016lip} to input into all classifiers. The network is initialized with pre-trained weights from the ImageNet dataset and fine-tuned on TCD-TIMIT images resized into $224\times224$px and fed into the CNN through the convolutional layers. Each convolutional layer consists of banks of convolutional filters with no changes to the original ResNet parameters. Following each convolutional layer a pooling layer performs down-sampling and learns positional invariance in the features. The penultimate layer is a fully connected layer on all features and final is an average pooling layer which outputs the computed features which are a deep image feature. 


\subsection{Classification} \label{sec:classification}
All classifiers have parameters which need tuning for optimal classification. We have not set out to explore the effects of the various classifier parameters on performance, rather we compare each classifier with others so to reduce causes of variation, the `sane default' parameter values are used. These are either taken from the implementation library, or related work in the literature. The classifiers surveyed include both image-based classification, such as Convolutional Neural Networks (CNN) or Support Vector Machines (SVM), while others are sequence-based classification methods, such as Hidden Markov Models (HMM) and Long-Short-Term-Memory (LSTM). Both image and sequence approaches have been used in speaker-dependent lipreading literature. 

First is a Gaussian \textbf{Naive Bayes} classifier \cite{zhang2004optimality} where the likelihood for each feature is assumed to be Gaussian and the parameters estimated using maximum likelihood. Implemented in the \texttt{GaussianNB} class of the \texttt{scikit-learn} library with the out-of-core training method. The classifier is trained incrementally one speaker at a time. 

The \textbf{Random Forest} classifier is implemented using the \texttt{sklearn.ensemble.RandomForestClassifier} class with ten estimators. Split quality is measured using the Gini impurity. Specifically nodes are split if the split would cause a decrease in the Gini impurity greater than zero, otherwise they are marked as leaves. All features are considered when looking for the best split. Additionally all the nodes in the tree are expanded until either all leaves are pure or there are less than two samples at each leaf. 
\cite{terissi2014lip} used random forests in a robust conventional MLR system.

\textbf{Support vector machines} are implemented with \texttt{scikit-learn} using $C$-SVM (\texttt{SVC}) where $C$ is a regularization parameter ($=1.0$), and the radial basis function kernel is $\gamma = \frac{1}{\#features}$. The remaining parameters are the library defaults (see `mathematical formulation' section of the \texttt{scikit-learn} documentation). For multi-class classification the `one-against-one' approach \cite{knerr1990single} is used as per other speaker-dependent SVM lipreading systems \cite{shaikh2010lip}.

We implement single-Gaussian \textbf{Hidden Markov Models} (HMM) using the \texttt{hmmlearn} library. Each three-state HMM is trained for every phoneme and viseme. HMM parameters are consistent with the best parameter options from prior work \cite{982900}. Predictions are made by computing the log-likelihood for a given feature vector over all of the HMMs for the given unit and selecting the best scored unit as our prediction. Predictions are assembled into sequences and compared against the ground truth to score accuracy. HMM's are the best performing conventional machine learning classifier for MLR \cite{bear2016decoding}.

The end-to-end \textbf{Convolutional Neural Network (CNN)} classification is a ResNet50 CNN model pre-trained on the ImageNet dataset \cite{imagenet_cvpr09}. Classification is the final fully connected layer of our model architecture is followed by Softmax regression. Training is completed in mini-batches of size $60$ using back propagation and stochastic gradient descent (SGD). The learning rate of the SGD decreases from $0.1$ to $0.001$ and momentum per mini-batch is $0.9$. Re-training the initialised model is in $20$ epochs per fold. The most accurate CNN MLR to date is achieved by \cite{Chung2018}. 

\textbf{Long short-term memory (LSTM)} networks are built in the \texttt{Keras} framework. Feature sequences are resized to a constant length of $190$ frames per sequence so each sequence is either trimmed to fit or padded with silence. 
The LSTM structure is; features are fed into a fully-connected layer, followed by three LSTM layers, followed by another fully-connected layer, a dropout layer and the activation layer. All the fully-connected layers (except the final layer) are of size $512$ and all the dropout layers have dropout value of $0.5$. The LSTM is trained with RMSprop and an adaptively decreasing learning rate from $0.01$. Training was stopped after ten epochs if the validation loss did not improve. LSTMs work well for MLR as they learn language in addition to lip shape and motion \cite{shillingford2018large}.


\section{RESULTS AND DISCUSSION}
\label{sec:print}

\begin{figure*}[!ht]
\centering
\begin{tabular}{cc}
\includegraphics[width=.49\textwidth]{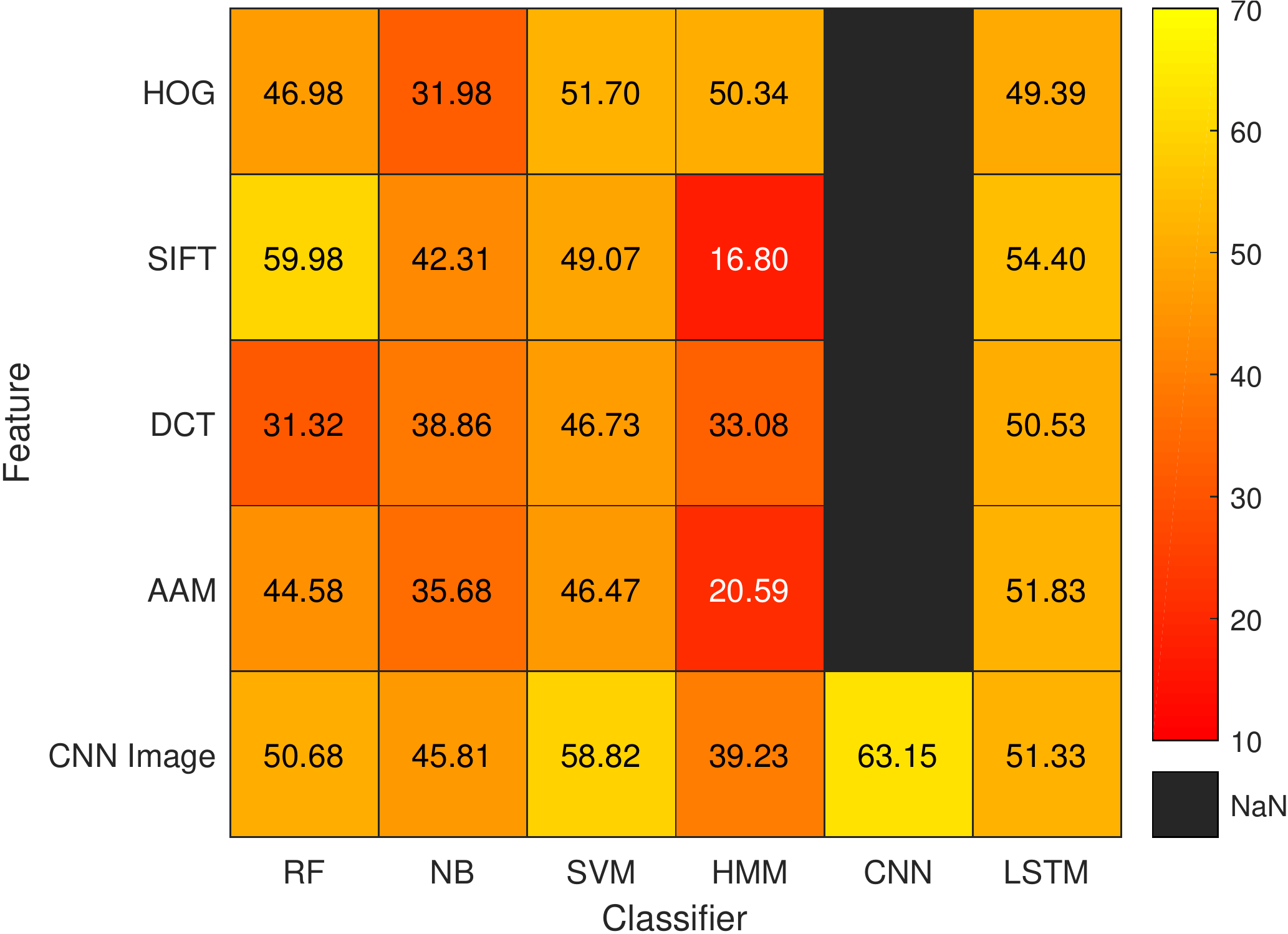} &
\includegraphics[width=.49\textwidth]{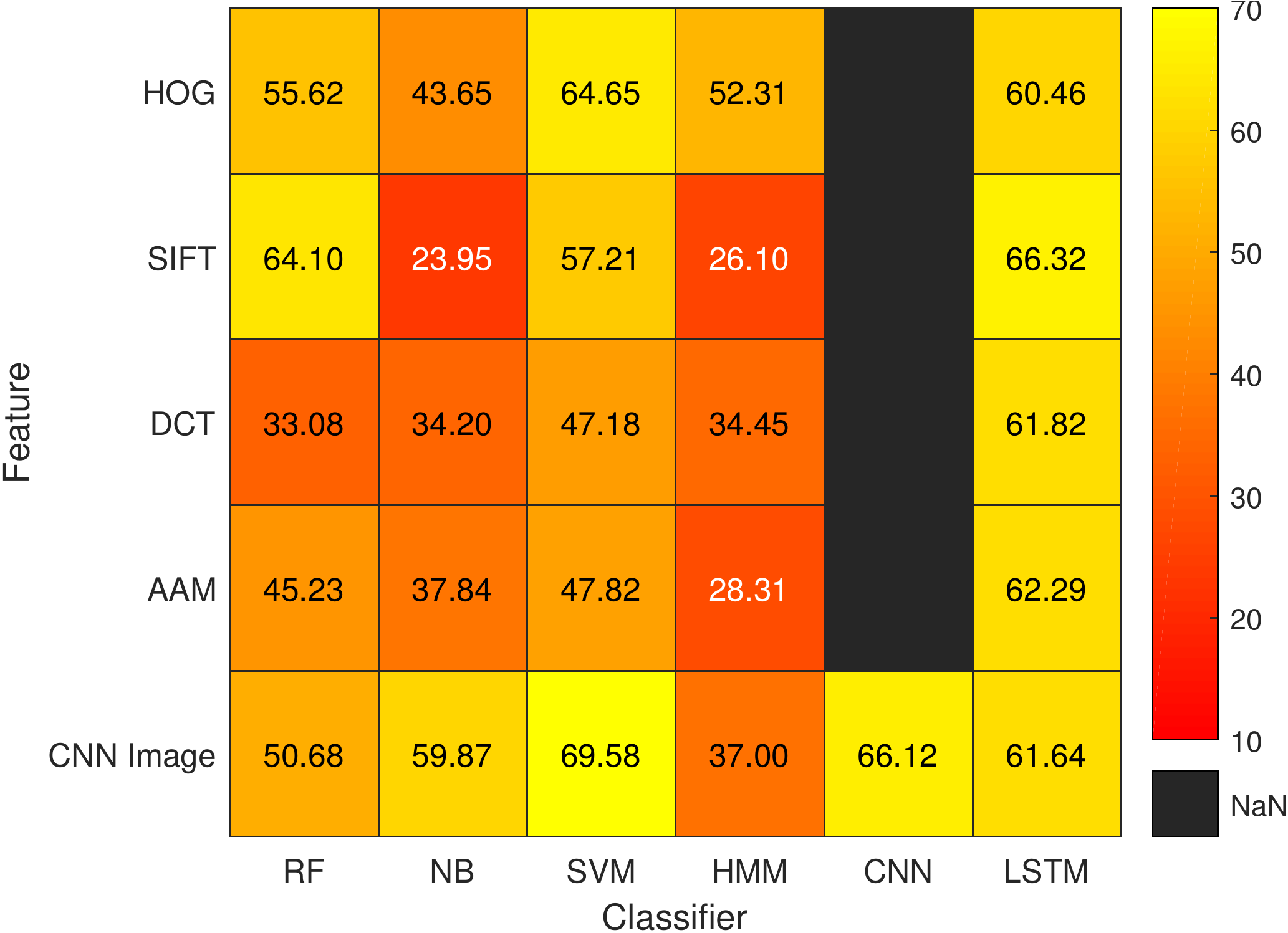} \\
a) Phoneme &  b) Viseme
\end{tabular}
\caption{Mean accuracy variation between lipreading system architectures on independent test speakers. Visemes mostly outperform phonemes, CNN features with SVM are most accurate for viseme classification, and phoneme classification is greatest with an end-to-end CNN.}
\label{fig:res}
\end{figure*}

We report top-$1$ accuracy values (plus/minus one standard error as a measure of significance) as the intraclass variation between visual speech classes is small but significant. We performed $5$-fold cross validation. Figure~\ref{fig:res} shows two heatmaps. The left heatmap shows lipreading accuracy with phonemes, the right heatmap reports visemes. For both heatmaps classifiers run along the $x$-axis and features run along the $y$-axis. Each cell contains the mean accuracy for that system architecture where the colour scale runs from red to yellow. Red is the least accurate. 

The phoneme heatmap (Figure~\ref{fig:res}-left), shows a accuracy high of $63.15\%$ with our end-to-end CNN architecture. The poorest phoneme lipreading system is built on SIFT features with an HMM classification scored $16.80\%$. The viseme heatmap (Figure~\ref{fig:res}-right) shows a accuracy high of $69.58\%$ with our Convolutional image-SVM system. The poorest lipreading system is built on SIFT features and a naive Bayes classifier with $23.95\%$ accuracy. 

It is interesting that the convolutional features with an SVM significantly outperformed all others, including the end-to-end trained CNN. We attribute this to the SVM being more suited to the lipreading problem than the softmax regression algorithm used in end-to-end CNNs. We had thought that the convolutional features with LSTM classification would outperform all others and we suggest that if trained end-to-end, that this accuracy would increase as the sequence rules of sentences would be learned into the model. AAM tracking is speaker-independent, had these been speaker-specific, the face tracking would be more accurate resulting in better features. However to retain the same variance in each speaker, we gained varying length feature vectors making training speaker-independent classifiers impossible.  


Our HMMs perform poorly overall despite previous literature demonstrating robust accuracy in conventional lipreading machines \cite{howell2016visual}. During experimenting we noted that some HMMs did not train due to insufficient occurrences of some phonemes or visemes. Therefore the error rates are negatively affected as these classes will be misclassified. Of the $38$ phonemes present in the ground truth of our dataset, $14$ did not have enough samples to train. This is despite the linguistic content of TCD-TIMIT having a good phonetic coverage (Section~\ref{sec:data}). 

It is unsurprising to see that deep learning methods outperform conventional approaches as the generalization learned by deep learning algorithms demonstrates a robustness to independent speakers if one assumes a similar speaker is in the training set. As expected this accuracy is lower than previous literature results which mask independent results with dependent test speakers (from Section~\ref{sec:back}, \cite{son2017lip} scored $76.2\%$). This is evidence that even with deep learning, speaker-independence is a problem with a significant affect on accuracy. 

When choosing between visemes or phonemes visemes are mostly more accurate than phonemes but not significantly so. The most accurate unit choice varies by system architecture. Based on these results unit choice decisions should be subject to not only the size of the dataset available to a researcher, but also the classification method to be implemented. 

\begin{figure*}[!]
\centering
\includegraphics[width=.49\textwidth]{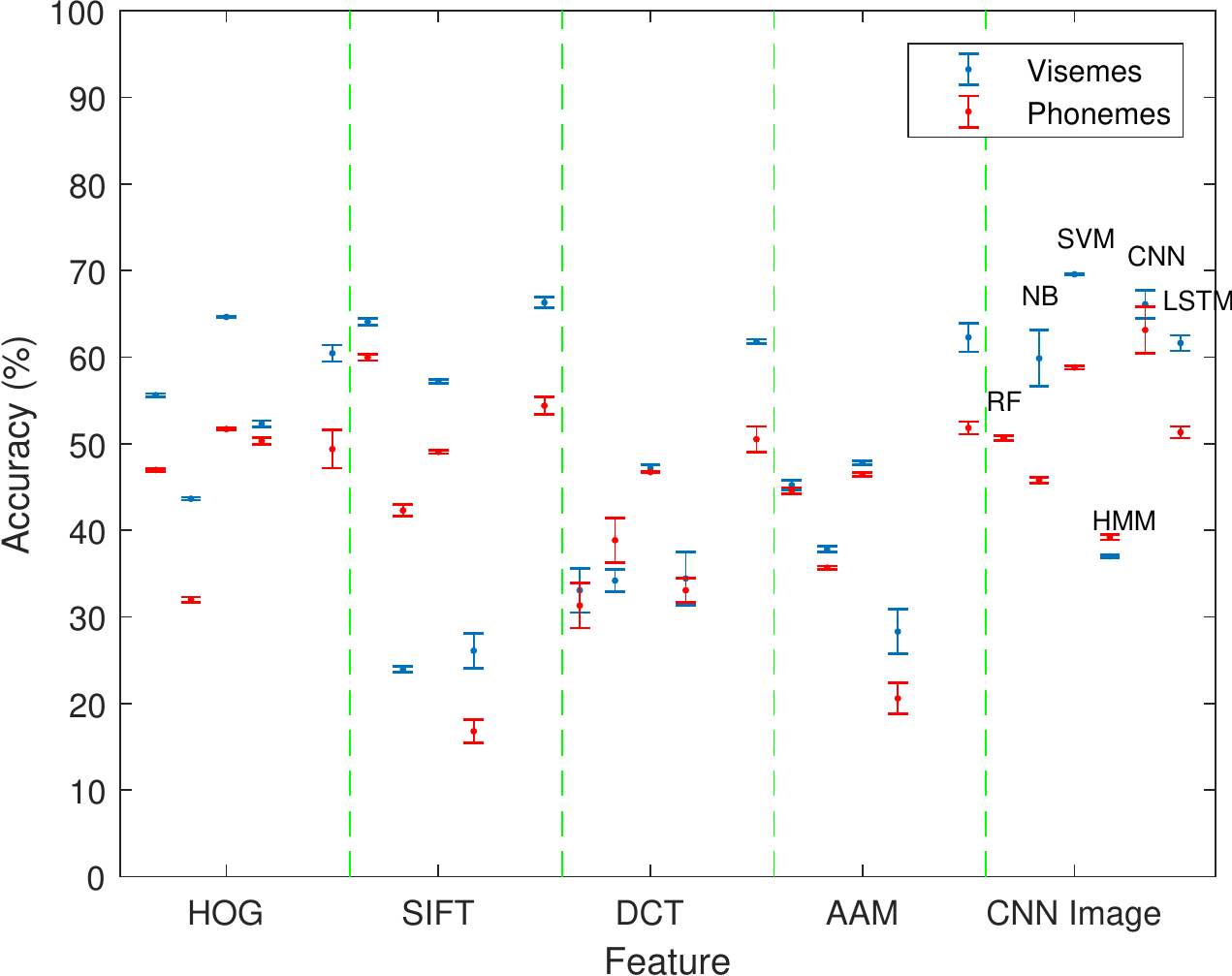} 
\includegraphics[width=.49\textwidth]{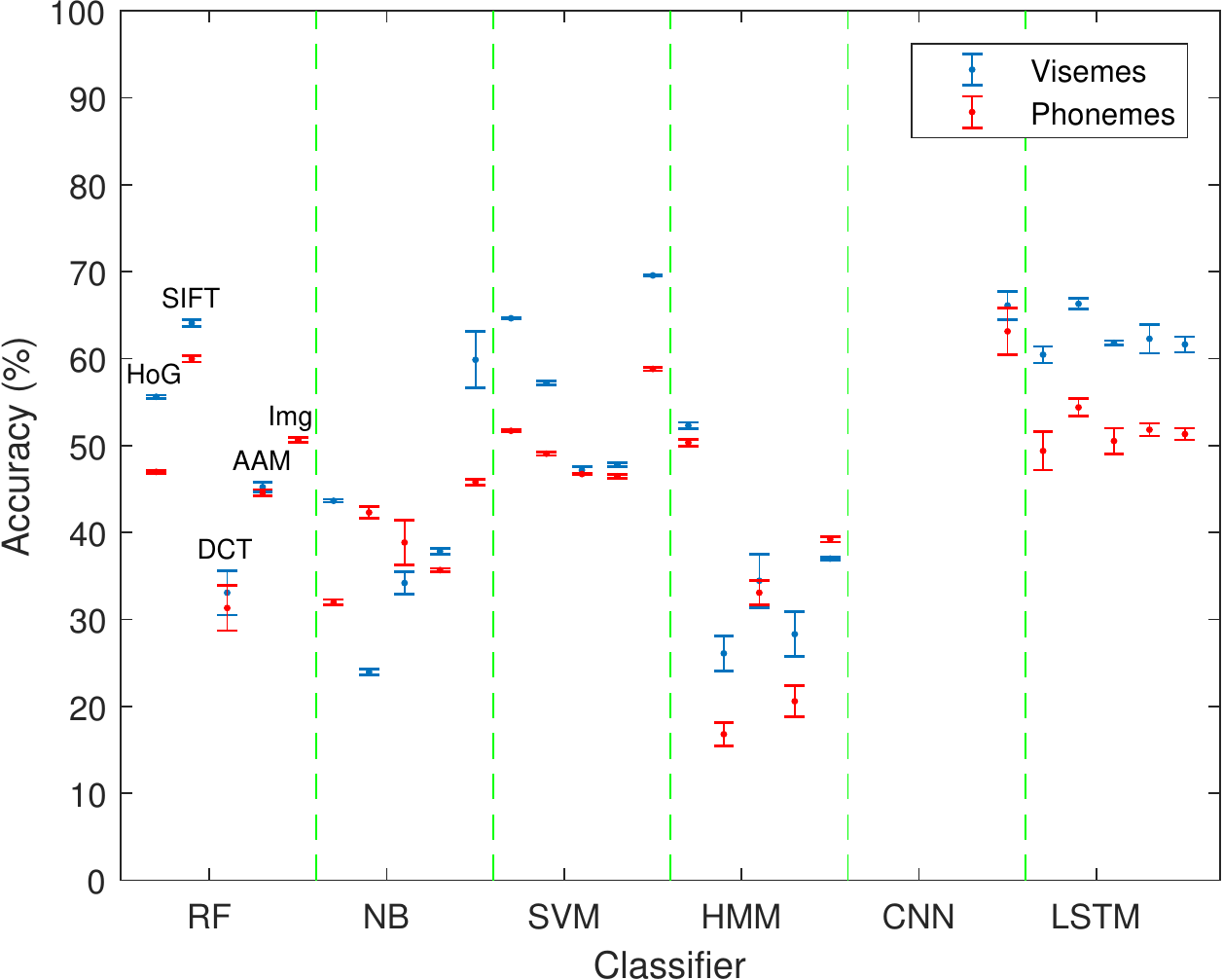} 
\caption{5-fold mean accuracy showing significant difference of $\pm$ one standard error, error bars.} 
\label{fig:play}
\end{figure*}

In Figure~\ref{fig:play} all scores are five-fold mean accuracy with error bars of $\pm$ one standard error. Figure~\ref{fig:play}-left lists our features along the $x$-axis for comparison and Figure~\ref{fig:play}-right lists our classification methods along the $x$-axis. Both visemes (in blue) and phonemes (in red) are presented.  On the right side of Figure~\ref{fig:play}-left we plot six points for the six classifiers as labelled. They are plotted in the same left-to-right order as the $x$-axis of Figure~\ref{fig:play}-right. This data presentation is consistent across the rest of Figure~\ref{fig:play}-left with each feature represented in its own section of the $x$-axis. Sections showing features HoG, SIFT, DCT, and AAM have spaces in the fifth position as the CNN features are only used in architectures for end-to-end CNN and with an LSTM. 
The HoG features are significantly affected by the classifier they are paired with in an Lipreading system showing a lack of robustness. The best HoG result is with SVM classifier and viseme units ($64.65\%$). For HoG features, visemes always outperform phonemes by at least $3\%$. Accuracies across the other features are significantly affected by the classifier pairing shown by many non-overlapping error bars. This suggests the classifier choice is more important than the feature choice (supported by Figure~\ref{fig:play}-right which shows much less variation in each section and more overlap of error bars).  As seen in the heatmaps (Figure~\ref{fig:res}), the best performance is achieved with Image-SVM-Viseme combination, but in Figure~\ref{fig:play}-left the error bar does not overlap with any other system architecture making it the current speaker-independent lipreading state-of-the-art benchmark. 

\section{CONCLUSIONS}
\label{sec:conclusions}
This paper surveys have undertaken a complete and systematic series of experiments to establish a benchmark for speaker-independent MLR. Using a consistent dataset, and cross-fold validation this work shows that a system based on image-based (convolutional) features and an SVM classifier achieves the most accurate lipreading for independent test speakers. The takeaway message of this work is that speaker-independent lipreading is an essential requirement for future lipreading machines for adoption in the real world and we have presented the next benchmarks needed for achieving this.

We have shown that in designing a speaker-independent lipreading machine the classifier choice has greater importance that the feature representation.
Overall the deep learning approaches; the CNN and the LSTM, are the most robust classifiers independent of the feature input, even though they did not win out over this survey. 

Speaker-dependent lipreading machines are reporting robust results. As research progresses to address challenges of lipreading in the real world, such as speaker-independence, this benchmark serves as a baseline for increasing the robustness of lipreading from the lab to the outside world. These results will serve our research community as a stepping stone towards achieving real-world applications that can lipread all speakers. 
\bibliographystyle{splncs}
\bibliography{refs}

\end{document}